\documentclass[JBR]{jbr_author} 

\title{Active Choice of Teachers, Learning Strategies and Goals for a Socially Guided Intrinsic Motivation Learner}

\articletype{Editorial}

\usepackage{subfig}

\usepackage{amsfonts}

\usepackage{fancyhdr} 
\usepackage{algorithmic}

\makeatletter
\newcounter{algorithmbis}
\renewcommand{\thealgorithmbis}{\thesection.\arabic{algorithmbis}}
\def\algorithmbis{\@ifnextchar[{\@algorithmbisa}{\@algorithmbisb}}
\def\@algorithmbisa[#1]{%
  \refstepcounter{algorithmbis}
  \trivlist
  \leftmargin\z@
  \itemindent\z@
  \labelsep\z@
  \item[\parbox{\linewidth}{%
    \hrule
    \hrule
    \noindent\strut\textbf{Algorithm \thealgorithmbis} #1
    \hrule
  }]\hfil\vskip0em%
}
\def\@algorithmbisb{\@algorithmbisa[]}

\makeatother

\author{Sao Mai Nguyen\inst{1} \email{nguyensmai at gmail.com},
        Pierre-Yves Oudeyer\inst{1} \email{pierre-yves.oudeyer at inria.fr}}

\institute{
     \inst{1} Flowers Team, INRIA and ENSTA ParisTech, France,\\
200 avenue de la Vieille Tour , 33 405 Talence Cedex, France
               }

\abstract{
We present an active learning architecture that allows a robot to actively learn which data collection strategy is most efficient for acquiring motor skills to achieve multiple outcomes, and generalise over its experience to achieve new outcomes. The robot explores its environment both via interactive learning and  goal-babbling. It learns at the same time when, who and what to actively imitate from several available teachers, and learns when not to use social guidance  but use active goal-oriented self-exploration. This is formalised in the framework of life-long strategic learning.

The proposed architecture, called Socially Guided Intrinsic Motivation with Active Choice of Teacher and Strategy (SGIM-ACTS), relies on hierarchical active decisions of what and how to learn driven by empirical evaluation of learning progress for each learning strategy. We illustrate with an experiment where a simulated robot learns to control its arm for realising two kinds of different outcomes. It has to choose actively and hierarchically at each learning episode: 1) {what to learn:} which outcome is most interesting to select as a goal to focus on for goal-directed exploration; 2) {how to learn:} which data collection strategy to use among self-exploration, mimicry and emulation; 3) once he has decided {when and what to imitate} by choosing mimicry or emulation, then he has to choose {who to imitate}, from a set of different teachers. We show that SGIM-ACTS learns significantly more efficiently than using single learning strategies, and coherently selects the best strategy with respect to the chosen outcome, taking advantage of the available teachers (with different levels of skills).}

\keywords{strategic learner \*\  imitation learning \*\ mimicry \*\ emulation \*\ artificial curiosity \*\ intrinsic motivation \*\ interactive learner \*\ active learning \*\ goal babbling \*\ robot skill learning}

\begin{document}
\maketitle
\thispagestyle{empty}
\thispagestyle{fancy}
\lhead{}
\chead{
\texttt{\scriptsize{S. M. Nguyen and P.-Y. Oudeyer. Active Choice of Teachers, Learning Strategies and Goals for a Socially Guided Intrinsic Motivation Learner,\\
 in Paladyn Journal of Behavioral Robotics,  September 2012, Volume 3, Issue 3, pp 136-146 . }}
\vspace{0pt}}
\rhead{}
%\cfoot{}

%% ###################################################################

 \begin{figure}[!h]
\centering
\includegraphics[width=0.35\textwidth]{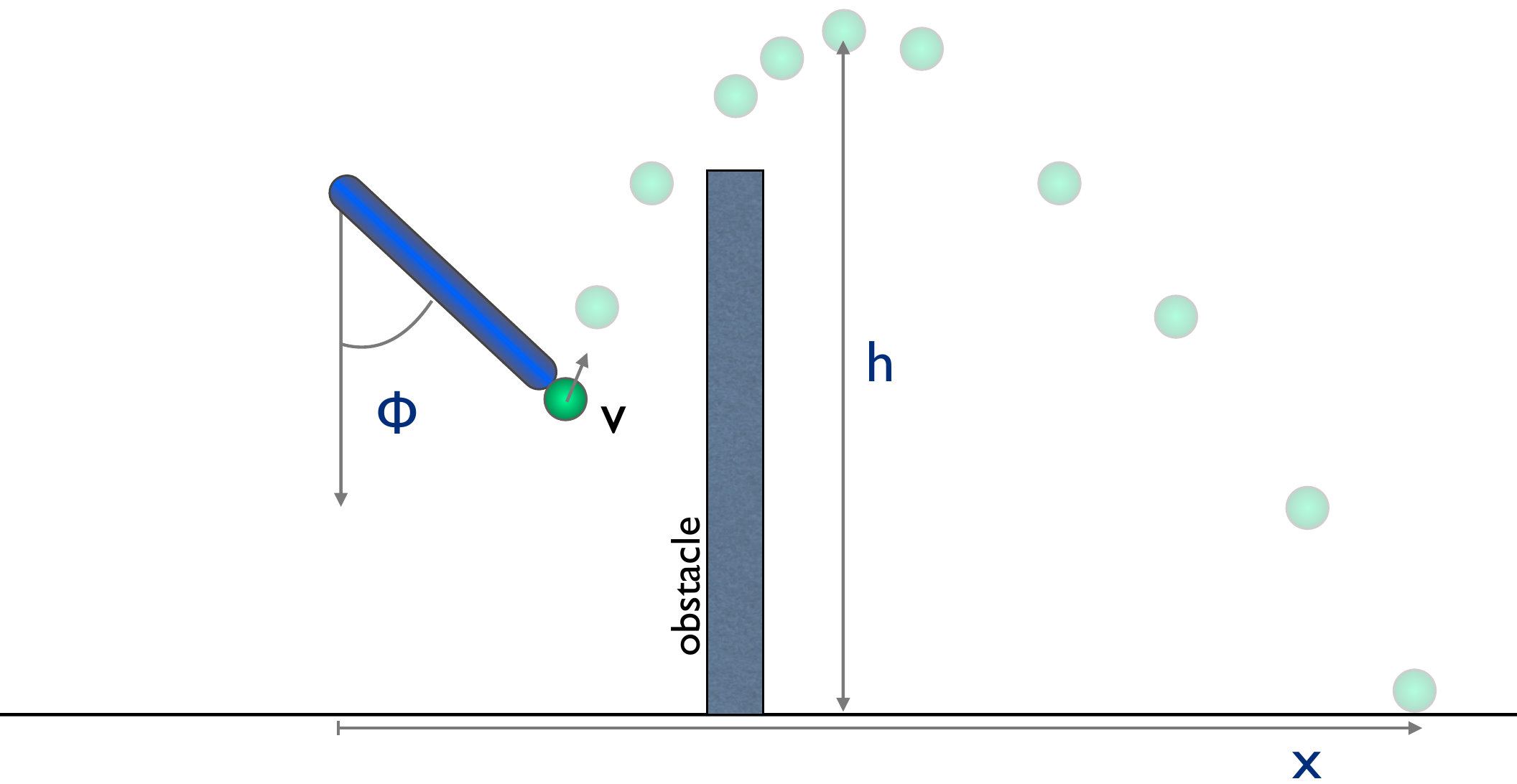}
\vspace{-0.6cm}
\caption{{\footnotesize 
%Experimental Setup: %An arm can throw a ball at different height and distances or point to different directions. 
An arm, described by its angle $\phi$, is controlled by  a motor primitive with 14 continuous parameters (taking bounded values) that determine the evolution of its acceleration $\ddot{\phi}$ . A ball is held by the arm and then released at the end of the motion. The objective of the robot is to learn the mapping between the parameters of the motor primitive and two types of outcomes he can produce: a ball thrown at distance x and height h, or a ball placed at the arm tip at angle $\phi$ with velocity smaller than $|v_{max}|$.
}
}
\vspace{-0.6cm}
\label{ExperimentalSetup}
\end{figure}

\section{Strategic Active Learning for Life-Long Acquisition of Multiple Skills}
\label{Intro}

Life-long learning by robots to acquire multiple skills in unstructured environments poses challenges of not only predicting the consequences or outcomes of their actions on the environment, but also learning the causal effectiveness of their actions for varied outcomes. The set of outcomes can be in large and high-dimensional sensorimotor spaces, while the physical embedding of robots  allows only limited time for collecting training data.  The learning agent has to decide for instance in which order he should focus on learning how to achieve the different outcomes, how much time he can spend to learn to achieve an outcome or which data collection strategy to use for learning to achieve a given outcome.
 
\subsection{Active Learning for Producing Varied Outcomes  with Multiple Data Collection Strategies}

These questions can be formalised under the notion of strategic learning \cite{Lopes2012ICDLE}.

 One perspective is learning to achieve varied outcomes. It aims at selecting which outcome to spend time on. A typical classification was proposed in \cite{Reichart2008,Qi2008CVPR} where active learning methods improved the overall quality of the learning. In sequential problems as in robotics, producing an outcome has been modelled as a local predictive forward model \cite{Oudeyer2007ITEC}, an option \cite{Barto2004IICDL}, or a region in a parameterised goal/option space \cite{Baranes2013RAS}. In these works each sampling of an outcome entails a cost. The learning agent has to decide which outcome to explore/observe next.  However most studies using this perspective do not consider several strategies.

Another perspective is learning how to learn, by making explicit the choice and dependence of the learning performance on the method. For instance, \cite{Baram2004JMLR} selects among different learning strategies depending on the results for different outcomes. However most studies using this perspective consider a single outcome.

Indeed, these works have not addressed the learning of both how to learn and what to learn, to select at the same time which outcome to spend time on, and which learning method to use. Only \cite{Lopes2012ICDLE} studies the framework of these questions, and only examined a toy example with discrete and finite number of states, outcomes and strategies. In initial work to address learning  for varied outcomes with multiple methods, we proposed the Socially Guided Intrinsic Motivation by Demonstration (SGIM-D) algorithm which uses both: 
\begin{itemize}
\item socially  guided exploration, especially programming by demonstration \cite{Billard2007RobotProgrammingby}, and 
\item intrinsically motivated exploration, which are active learning algorithms based on measures of the evolution of the learning performance \cite{Oudeyer2007FN}
\end{itemize}
to reach goals in a continuous outcome space, in the case of a complex and continuous environment. High-dimensional environments can be handled by SGIM-D, designed  for multiple outcomes in a continuous outcome space. In \cite{Nguyen2011IICDL}, SGIM-D learned to manipulate a fishing rod with a 6-dof arm, i.e. to place the float on the surface of the water, which is described as a 2d continuous outcome space. The robotic arm was controlled by a motor primitive with 24 continuous parameters that determine the trajectory of its joint positions. The robot learned   which action $a$ to perform  for a given goal position on the surface of the water $y_g$, where the hook should reach when falling into the water. However, the outcomes considered belonged to only one type of outcomes. Moreover, although SGIM-D has 2 learning strategies, it is a passive learner which only imitates when the teacher decides to give a demonstration. SGIM-D does not learn which method enables it to perform best.

 In this paper, we address these two limitations. We study how a learning agent can achieve varied outcomes in structured continuous outcome spaces, even with outcomes of different types, and how he can learn for those various outcomes which strategy to adopt among 1) active self-exploration, 2) emulation of a teacher actively selected among available teachers, 3) mimicry of an actively selected teacher. We propose an algorithm for actively choosing the appropriate strategy, among several strategies.

\subsection{Formalisation}

 \begin{figure}
  \centering
  \includegraphics[width=0.5\textwidth]{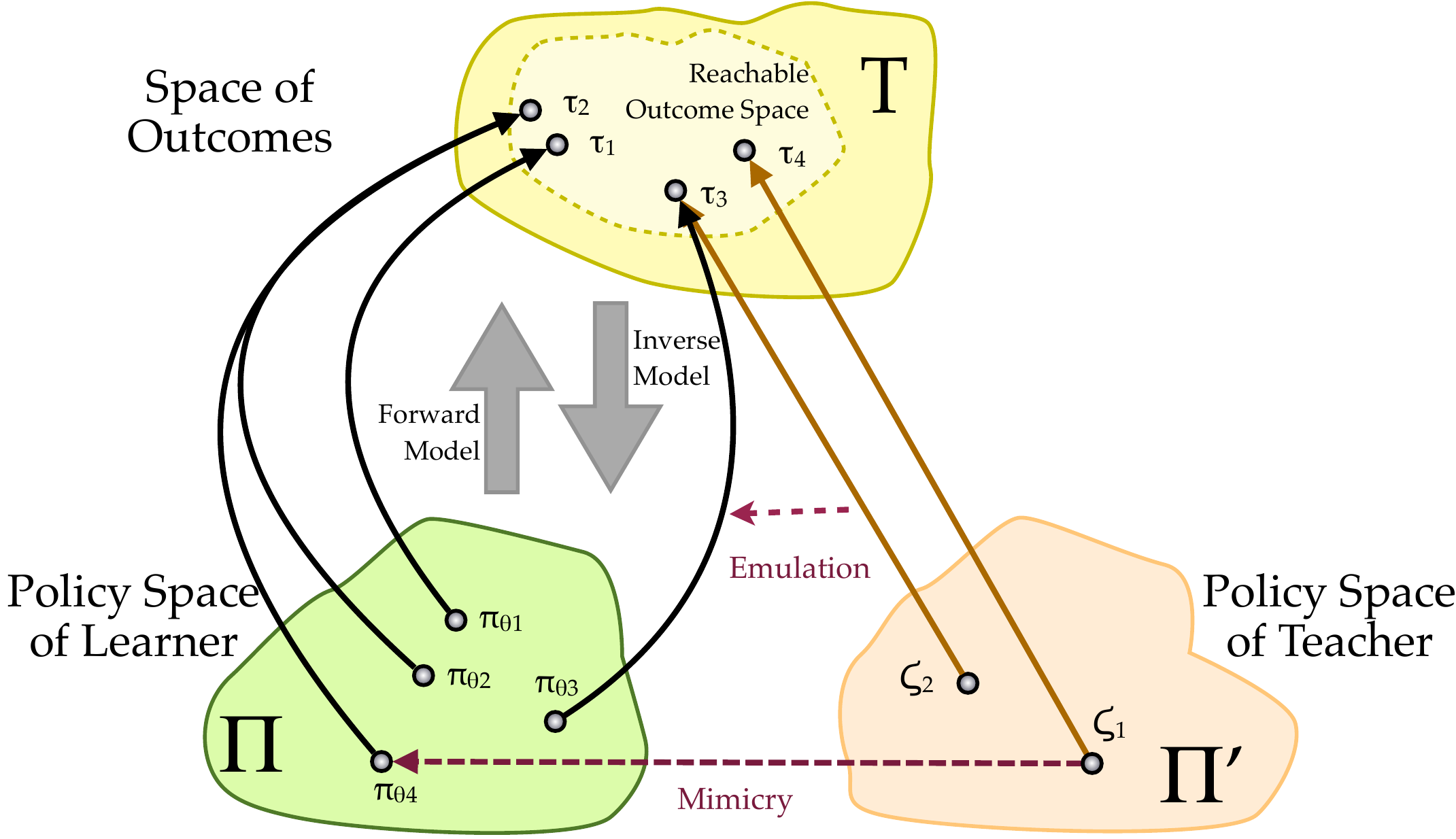}

\caption{
{\footnotesize 
Representation of the problem. The environment can evolve to an outcome state $\tau$ by means of the learner's policy of parameter $\theta$ or the teacher's  actions $\zeta$. The learner and the teacher have a priori different policy spaces.  The learner estimates $L^{-1}: T \mapsto \Pi $. By emulation or mimicry, the learner can take advantage of the demonstrations  ($\zeta$,$\tau_d$) of the teacher to improve its estimation $L^{-1}$.}
}
\vspace{-0.9cm}
\label{DataFlow}
\end{figure}

 Let us consider an agent learning motor skills, i.e. the mapping between an outcome space and a policy space. As an illustration, let us imagine the agent learning how to play tennis, He maps how the ball behaves (outcome) with respect to the movement of his racket (policy). He thus learns a forward model $M$ to predict where the ball bounces given the movement of his racket. More importantly, he builds an inverse model $L^{-1}$ to control his racket in order to make the ball bounce at a desired position. A good player knows which outcomes are feasible and knows at least one policy to produce any possible outcome: he can place the ball anywhere on the court. Ideally, he builds an inverse model $L^{-1}$ such that $M(L^{-1})$ is identity. 
 
More formally, we define an outcome space which may comprise of outcomes of different types and different dimensionalities. For tennis, outcomes can be the bouncing positions, spin angles ... We only assume that they can be parameterised by parameters $\tau \in T$ and that we can define a distance measure $J$ on $T \times T$.
A policy $\pi_\theta$  is described by motor primitives parameterised by $\theta \in \Pi$. Its outcome is $M(\theta)$, where  the mapping $M:\Pi \to T$ describes the environment. For the tennis player, the policy controls the movement of his arm and racket and $M$ represents the physical equations for the ball trajectory.
    The performance of a policy $\pi_\theta$ at completing an outcome $\tau$ is measured by the distance between $\tau$ and the outcome of $\pi_\theta$: $J(\tau, M(\theta))$.
 %A policy $\pi_\theta$ parameterised by $\theta \in \Pi$ gives the probability that action $a$ is the right action to perform :
%   \begin{align}
% \pi_\theta: & A & \to & [0,1]  \nonumber \\
%         & a  & \mapsto &\pi_{\theta}(a) \nonumber
% \end{align}

The agent focuses on learning the inverse model and builds its estimate $L^{-1}: T \to \Pi $.  We note that $M^{-1}$, the inverse of $M$ might not be a function as $M$ might be redundant, whereas our learner builds a function $L^{-1}$ that finds at least one adequate policy to complete every outcome $\tau$.  
In sum, it endeavours to minimise with respect to $L^{-1}$ : 
  \begin {eqnarray}
I =   \int_{\tau \in T} P(\tau)J(\tau,M(L^{-1}(\tau))) d\tau
  \end{eqnarray}
where $P(\tau)$ is a probability density distribution over $T$. A priori unknown to the learner, $P(\tau)$ can describe the probability of $\tau$ occurring or the reachable space or a region of interest.

    We assume that $T$ can be partitioned into subspaces where the outcomes are related, and in these subspaces our parametrisation allows a smooth variation of $\tau \mapsto J(\tau,M(\theta)), \forall \theta$ with respect to $\tau$  most of the time.  This partition, initially unknown to the agent, needs to be learned.
   
   Note that we have described our method without specifying a particular choice of policy representation, learning algorithm, action or outcome space properties. These designs can indeed be decided according to the application at hand.
 In particular, outcomes can be of different types and dimensionalities.  In this case, we note $T_i$ the subspaces of $T$ corresponding to the different types of outcome and $T= \cup T_i$.

\subsection{Our Approach}

To solve the problem formalised above, we propose a system, called Socially Guided Intrinsic Motivation with Active Choice of Teacher and Strategy ({\bf SGIM-ACTS}) that allows an online interactive learning of inverse models in continuous high-dimensional robotic sensorimotor spaces with multiple teachers, and learning strategies. SGIM-ACTS learns various outcomes with different types of outcomes, and generalises from sampled data to continuous sets of outcomes.

Technically, we adopt a method of generalisation of policies for new outcomes similar to \cite{Kober2012AR,Silva20122ICMLI2}. 
Whereas in their approaches  the algorithms use a pool of examples given by the teacher preset from the beginning of the experiment to learn outcomes specified by the engineer of the robot, in a batch learning method;  in our case,  the SGIM-ACTS algorithm decides by itself which outcomes it needs to learn more to better generalise for the whole outcome space, like in \cite{Oudeyer2007ITEC,Barto2004IICDL,Baranes2013RAS}.  Moreover, SGIM-ACTS actively requests the teacher's demonstrations online, by choosing online the best learning strategy, similarly to \cite{Baram2004JMLR}, except that we do not learn with a discrete outcome space for a classification problem, but with a continuous outcome space. SGIM-ACTS also interacts with several teachers and uses several social learning methods, in an interactive learning approach.

Our active learning approach is inspired by:
\begin{itemize}
\item intrinsic motivation in psychology \cite{Ryan2000CEP} which triggers spontaneous exploration and curiosity in humans, which recently led to novel robotic and machine active learning methods which outperform traditional active learning methods \cite{Baranes2013RAS,Lopes2012NIPSN}
\item teleological learning \cite{Csibra2003PTRSLSBS} which considers actions as goal-oriented, and recently led to efficient goal babbling methods in robotics \cite{Baranes2013RAS,Rolf2012PH2WDRDRYHCA}
\item psychological theories for socially guided learning \cite{Call2002Threesourcesof,Dautenhahn2002,Tomasello2007DS}, as detailed in the next section. 
\end{itemize}

After this formal description of our approach, we analyse our point of view on social guidance  in section \ref{Soc}. Then, we detail the proposed algorithm SGIM-ACTS in section \ref{Algo}, before testing it on a problem to learn how to throw and place a ball (fig. \ref{ExperimentalSetup}) in section \ref{Exp}.

\section{Social Guidance}
\label{Soc}

\subsection{Interactive Learning}

%Indeed, the SGIM-D learner is passive with respect to the social interaction and the teacher, as he imitates only and automatically when the teacher gives a demonstration. Moreover, he only interacts with a single expert user, and does not optimise the timing of the interactions with the teacher.

An interactive learner who not only listens to the teacher, but  actively requests for the information  it needs and when it needs help, has been shown to be a fundamental aspect of social learning \cite{Chernova2009JAIR,Thomaz2006,Nicolescu2003PSIJCAAMS}.
Under the interactive learning approach, the robot can combine programming by demonstration, learning by exploration and tutor guidance. Several works in interactive learning have considered extra reinforcement signals \cite{Thomaz2008CS}, action requests \cite{Grollman2010IRS,Lopes2009ECML} or disambiguation among actions \cite{Chernova2009JAIR}. In \cite{Cakmak2010AMDIT} the comparison of a robot that has the option to ask the user for feedback, to the passive robot, shows a better accuracy and fewer demonstrations.
Therefore, requesting demonstrations  when it is needed can lessen the dependence on the teacher and reduce the quantity of the demonstrations required.  This approach is the most beneficial to the learner, for the information arrives as it needs it, and to the teacher who no longer needs to monitor the learning process. 

For an agent learning motor skills, i.e. the mapping between policies and outcomes, let us examine the type of social guidance that a learner can get  as reviewed in \citep{Argall2009RAS,Billard2007RobotProgrammingby,Schaal2003PTRSLSBS,Lopes2009AbstractionLevelsfor} with respect to: what, how, when and who \citep{Dautenhahn2002}. In this section, we note $si_H$ the information flow from the human to the robot. 

\subsection{What?}
Let us  examine the target of the information given by the teacher, or mathematically speaking, the space on which he operates. This can be either the policy or outcome spaces, or combinations of them.

\subsubsection{Policy Space}
Many social learning studies target the policy parameter space $\Pi$. For instance, in programming by demonstration (LbD), $si_H$ shows the right policy to perform in order to reach a given goal. As an illustration, when teaching how to play tennis, your coach could show you how to hit a backhand by a demonstration, or even by taking your hand and directing your movement. This approach relates to two levels of social learning: \textit{mimicry}, in which the learner copies the policies of others without an appreciation of their purpose, and \textit{imitation}, in which the learner reproduces the policies and the changes in the environment, as formalised in \citep{Lopes2009AbstractionLevelsfor,Call2002Threesourcesof,Whiten2000CS}.
  %Action mimicking  \cite{Cakmak2009IISRHIC} or  human teachers' corrections \cite{Chernova2009JAIR} have been implemented in robotic systems. 
%   The policies demonstrated can be mimicked faithfully \citep{Cakmak2009IISRHIC}, be saved as corrections for the current situation \citep{Chernova2009JAIR}, form an initial dataset on which to build upon more complicated behaviour\citep{Argall2008,Argall2011RAS}, or indicate a locality from where to search for an optimum \citep{Peters2008NN}.   %It can also take the form of semi-supervised classification: the Confidence-Based Autonomy algorithm uses a classifier that associates the context to the action, and the human teaches the right action for a given context \cite{Chernova2009JAIR}.
%  The information can be a trajectory or policy\citep{Peters2008NN},  high-level instructions\citep{Thomaz2006} or high-level advice\citep{Argall2008,Argall2011RAS}.  It can pertain to the entire policy, or only a part of it \citep{Argall2008,Argall2011RAS,Nicolescu2003PSIJCAAMS,Thomaz2006}.
The literature often considers that targeting the policy space is the most directive and efficient method. However, it relies on the human teacher's expertise, which bears limitations such as ambiguity, imprecision, under-optimality or  the correspondence problem. % Furthermore, the interaction is more effective at correcting visited situations, than exploring undemonstrated areas of $T$.

\subsubsection{Outcome Space}
The second kind of information is about possible outcomes $\tau \in T$, and is related to goal-directed exploration, where the learner focuses on discovering different outcomes instead of different ways of entailing the same outcome.  Psychologically speaking, this case pertains to the \textit{emulation} level of social learning, where the observer witnesses someone produce a result on an object, but then employs his own policy repertoire to reproduce the result, as formalised in \citep{Lopes2009AbstractionLevelsfor,Call2002Threesourcesof,Whiten2000CS,Nehaniv2007}. %The learner focuses on the observed effects of the actions of the demonstrator, possibly reaching these after adjusting his action choice \cite{}. A teacher can therefore indicate a new possible outcome/goal the learner should reach. 
During our tennis training, your coach could ask you to hit with the ball the right corner of the court, wherever you received the ball, whichever shot you use.
%. Independently from how or where you received the ball, and without any instructions on whether you need to hit with an overheard smash, or forehand or a backhand, all you need to train for is to make the ball reach its goal on the right corner of the court. 
Goal-directed approaches  allow the teacher to reset goal outcomes \citep{Argall2008}, to request the execution of outcomes \citep{Thomaz2006} or to label outcomes \citep{Thomaz2006, Thomaz2008CS}.  The learner can infer from the demonstrations the goal outcome by positional and force profiles to iron and open doors \citep{Kormushev2011AR}, or by using inverse reinforcement learning \citep{Lopes2011IICDL}.
This approach is essential to learn multiple outcomes, and all the more interesting as it is inspired by psychological behaviours \citep{Whiten2000CS,Tomasello2007DS,Csibra2003PTRSLSBS}. The drawback is that the learning needs the actions repertoire to be large enough to be used to reach various goals, before it improves.

%\paragraph{Context and Action Spaces:}
%Numerous methods target the context and action spaces at the same time. This is particularly the case of methods based on reinforcement learning. In Sophie's kitchen experiment, human teachers were asked to reward the robot on the $(c,a)$ pair \cite{Thomaz2006}, indicating the desirability of the current behaviour.

As we want the learner to accomplish not only a single outcome but to be efficient on a large variety of goals, we choose to bootstrap its learning with information targeting the outcome space. Furthermore, we also want the learning process to benefit from the social interaction early. So that the learner builds its action repertoire quickly, we choose to target the policy parameter space $\Pi$ too.

\subsection{When?}
The timing of the interaction varies with respect to its general activity during the whole learning process.
The rhythm of social interaction varies considerably among studies of social learning:

\begin{itemize}
\item{At a fixed frequency:} In classical imitation learning, the learner uses a demonstration to improve its learning at every policy it performs \citep{Argall2008,Argall2011RAS,Cakmak2009IISRHIC}. 
This solution is ill-adapted to the teacher's availability or the needs of the learner who requires more support in difficult situations. %Though, this continuous interaction allows steady bootstrapping of the learning and adaptation to changing environments.

\item{Beginning of learning:} A limited number of examples are given to initialise the learning, as a basic behaviours repertoire \citep{Argall2008,Argall2011RAS}, or  a sample behaviour to be optimised \citep{Peters2008NN,Kormushev2010IROS}. The learner is endowed with some basic competence before self-exploration. Nevertheless, if the interactions are restricted to the beginning, the learner could face difficulties adapting to changes in the environment.

\item{At the teacher's initiative:} The teacher alone decides when he interacts with the robot \citep{Thomaz2006}, by for instance giving corrections when seeing errors \citep{Koenig2010NN,Cakmak2010AMDIT}. Nevertheless, it still is time consuming as he needs to monitor the robot's errors to give adequate information to the learner.

\item{At the learner's initiative:} The interactive learner can request for the teacher's help in an ambiguous \citep{Chernova2009JAIR,Cakmak2010AMDIT}  or unknown \citep{Thomaz2006} situation, or only reproduces the observations when the observed outcome matches its goal  during goal-based imitation or mimicking \citep{Cakmak2009IISRHIC}. This approach is the most beneficial to the learner, for the information arrives as it needs them, and the teacher needs not monitor the process. 
\end{itemize}
 
 These 4 types can be classified into 2 larger groups: 
 \begin{itemize}
 \item batch learning, where the data provided to the learner is decided before the learning phase, and is given independently of the learning progress, generally in the beginning of the learning phase.
 \item interactive learning, where  the user interacts with the incrementally learning robot, either at the teacher's or the learner's initiative.
 \end{itemize}
 
 \subsection{Who?}
 While most social guidance studies only consider a single teacher, in natural environments, a household robot in reality interacts with several users. %And asking help from one or the other teacher is an important issue,  particularly when teachers have different levels of competence for different kinds of outcomes.
 Moreover, being able to request help to different experts is also an efficient way to address the problem of the reliability of the teacher. Imitation learning studies often rely on the quality of the demonstrations, whereas in reality a teacher can be performant for some outcomes but not for others. Demonstrations can be ambiguous, unsuccessful or suboptimal in certain areas. Like students who learn from different teachers who are experts in the different topics of a curriculum, a robot learner should be able to determine its best teacher for the different outcomes it wants to achieve.

 In this work, we consider the possibility of a learner to observe and imitate from several teachers, as much like a child in a natural environment would observe and imitate several adults in his surrounding throughout his development. In this case, choosing  whom to imitate, recognising who is the expert in the outcomes we need to make progress, constitutes an important strategy choice.

\subsection{Actively Learning When, Who and What to Imitate}
For the model and experiments presented below, our choice of social guidance among this listing of social learning is:
 \begin{itemize}
 \item{What:} We opted for an information flow targeting both policy and outcome spaces,  to enable the biggest progress for  the learner. It can imitate to reproduce either a demonstrated policy or outcome. Therefore, our learner can decide whether to \textit{mimic} and \textit{emulate} by learning what is the most interesting information.
 \item{When:} Interactive learning at the \textit{learner's initiative} seems the most natural interaction approach, the most efficient for learning and less costly for the teacher than if he would have to monitor the learner's progress to adapt his demonstrations. The robot has to learn when it is useful to imitate.
 \item{Who:}  Interactive learning where the learner can \textit{choose who} to interact with and to whom to ask for help, is an important strategy choice in learning.
 \end{itemize}

Thus, it learns to answer the four main questions of imitation learning: "what, how, when and who to imitate" \cite{Dautenhahn2002, Breazeal2002TCS} at the same time. We address  active learning  for varied outcomes with multiple strategies, multiple teachers, with a structured continuous outcome space (embedding sub-spaces with different properties). The strategies we consider are autonomous self-exploration, emulation and mimicking, by interactive learning with several teachers.
Hereafter we describe the design of our \textbf{SGIM-ACTS} (Socially Guided Intrinsic Motivation with Active Choice of Teacher and Strategy)  algorithm. Then we show through an illustration experiment that SGIM-ACTS efficiently learns  to realise  different types of outcomes  in continuous  outcome spaces, and it coherently selects the right teacher to learn from.

\section{Algorithm Description}
\label{Algo}

In this section, we describe the SGIM-ACTS architecture by giving a behavioural outline in section \ref{Outline}, before describing its general structure in section \ref{Hierarchical}. We then detail the different functions in sections \ref{Pol} and \ref{StratTask}. The overall architecture is summarised in Algorithm \ref{alg:PseudoCodeSGIMACTS} and is illustrated in fig. \ref{StructureSGIM} .

\begin{figure}
{\small
\input{PseudoCodeSGIMACTS.tex}
}
\vspace{-1cm}
\end{figure}

\subsection{Architecture Outline}
\label{Outline}

\begin{figure*}
\centering
\includegraphics[width=0.8\textwidth]{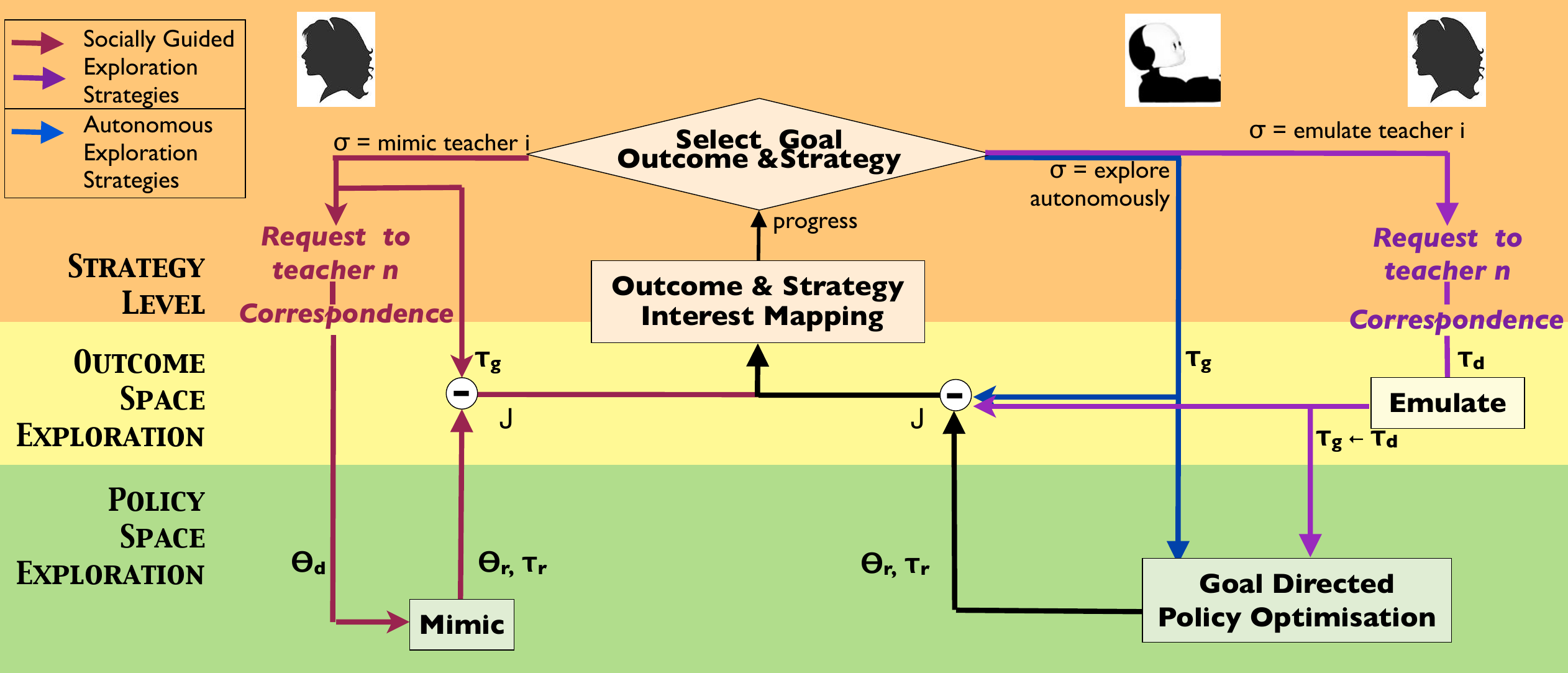}
\vspace{-0.5cm}
\caption{Time flow chart of SGIM-ACTS, which combines Intrinsic Motivation and Mimicking and Emulation into 3 layers that pertain to the strategy, the outcome space  and the policy space exploration respectively.}
\label{StructureSGIM}
\vspace{-0.5cm}
\end{figure*}

%Our learner  improves its estimation $L^{-1}$ to maximise $I = \int_{\tau} P(\tau)J(\tau,L^{-1}(\tau)) d\tau$ both by self-exploring the policy and outcome space and by asking for help to a teacher, who performs an observed trajectory $\zeta$ to achieve an outcome $\tau_d$. 
SGIM-ACTS is an architecture that merges intrinsically motivated self-exploration with interactive learning as socially guided exploration. In the latter case, a teacher performs an observed trajectory $\zeta$ which achieves an outcome $\tau_d$.
Note that the observed trajectory might be impossible for the learner to re-execute, and he can only approach it best with a policy $\pi_{\theta_d}$. 

The agent learns to achieve different types of outcomes by actively choosing which outcomes to focus on and set as goals, which data collection strategy to adopt and to which teacher to ask for help. It learns local inverse and forward models in complex, redundant and continuous spaces.

SGIM-ACTS learns by episodes during which it actively chooses simultaneously an outcome $\tau_g \in T$ to reach and a learning strategy  with a specific teacher  (cf. \ref{SectionSelectTaskStategy}). Its choice $\sigma$ is selected between : intrinsically motivated exploration, mimicry from teacher 1, emulation of  teacher 1,  mimicry from teacher 2, emulation of  teacher 2 .... 

In an episode under a mimicking strategy (fig. \ref{StructureSGIM}), our SGIM-ACTS learner actively self-generates a goal $\tau_g$ where its competence improvement is maximal (cf. \ref{SectionSelectTaskStategy}). The SGIM-ACTS learner explores preferentially goal outcomes easy to reach and where it makes progress the fastest. The selected teacher answers its request with a demonstration  $[\zeta_{d},\tau_{d}]$ to produce an outcome $\tau_{d}$ that is closest to $\tau_{g}$ (cf. \ref{SectionMimic}). The robot mimics the teacher to reproduce $\zeta_d$, for a fixed duration, by performing policies $\pi_\theta$ which are small variations of an approximation of $\zeta_d$.

In an episode under an emulation strategy (fig. \ref{StructureSGIM}), our SGIM-ACTS learner observes from the selected teacher a demonstration $[\zeta_{d},\tau_{d}]$.  It tries different policies using goal-directed optimisation algorithms to approach the observed outcome $\tau_d$, without taking into account the demonstrated policy  $\zeta_d$. It  re-uses and optimises its policy repertoire built through its past autonomous and socially guided explorations (cf. \ref{GoalDirectedPolicyOptimisation}).  The episode ends after a fixed duration.

In an episode under the intrinsic motivation strategy (fig. \ref{StructureSGIM}), it explores autonomously following the SAGG-RIAC algorithm  \cite{Baranes2013RAS}. It actively self-generates a goal $\tau_g$ where its competence improvement is maximal (cf. \ref{SectionSelectTaskStategy}), as in the mimicking strategy. Then, it explores which policy $\pi_\theta$ can achieve $\tau_g$ best. 
 It tries different policies to approach the self-determined outcome $\tau_g$, as in the emulation strategy (cf. \ref{GoalDirectedPolicyOptimisation}). The episode ends after a fixed duration. The intrinsic motivation  and emulation strategies differ mainly by the way the goal outcome is chosen.

An extensive study of the role of these different learning strategies can be found in \cite{Nguyen20122IISRHIC}. Thus the mimicry exploration increases the learner's policy repertoire on which to build up emulation and self-exploration, while biasing the policy space exploration. Demonstrations with structured policy sets, similar policy shapes, bias the policy space exploration to interesting subspaces, that allow the robot to overcome high-dimensionality and redundancy issues and interpolate to generalise in continuous outcome spaces.
With emulation learning, the teacher influences  the exploration of the outcome space. He can hinder the exploration of subspaces attracting the learner's attention to other subspaces. On the contrary, he can encourage their exploration by making demonstrations in those subspaces.
Self-exploration is essential to build up on these demonstrations to overcome correspondence problems and collect more data to acquire better precision according to the embodiment of the robot.

This behavioural description of SGIM-ACTS is followed in the next section by the description of its architecture.

\subsection{Hierarchical Structure}
\label{Hierarchical}

% It learns to achieve different types of outcomes by actively choosing which outcomes to focus on, and which learning strategy to adopt. It learns local inverse and forward models in complex, redundant and continuous spaces.
 SGIM-ACTS improves its estimation $L^{-1}$ to minimise $I = \int_{\tau} P(\tau)J(\tau,M(L^{-1}(\tau))) d\tau$ by exploring with the different strategies the outcome and policy spaces.
  Its architecture is separated into three levels:
\begin{itemize}
\item A \textit{Strategy Exploration} level which decides actively which learning strategy to use between intrinsic motivation, emulation and mimicry, and which teacher to ask for demonstrations  (\textit{Select Goal Outcome and Strategy}).  To motivate its choice, it maps $T$ in terms of interest level for each strategy (\textit{Outcome and Strategy Interest Mapping}) to keep track which strategy and which subspace of $T$ leads to the best learning progress.
\item An \textit{Outcome Space Exploration} level which minimises $I$ by exploring T. It decides actively which outcome $\tau_g$ to focus on, to minimise $J(\tau_g,M(L^{-1}(
\tau_g)))$, according to the adopted strategy. In the case of an emulation strategy, it sets the observed outcome of the demonstration $\tau_d$ as a goal. In the case of  mimicry and intrinsic motivation strategies, it  self-determines a goal $\tau_g$ selected by the \textit{Select  Goal Outcome and Strategy} function.
\item A \textit{Policy Space Exploration} level which explores the policy parameters space $\Pi$ to improve its estimation of $J$ and estimate the inverse mapping $L^{-1}(\tau_g)$. With the mimicry learning strategy, it mimics the demonstrated trajectory $\zeta_d$ by the chosen teacher to estimate $J$ around that locality (\textit{Mimicry}). With the emulation and autonomous exploration strategy, the \textit{Goal-Directed Policy Optimisation} function minimises $J(\tau_g,M(\theta))$ with respect to $\theta$. It attempts to reach the goals $\tau_g$ set by the Strategy and Outcome Space Exploration level, and gets a better estimate of $J$ that it can use later on to reach other goals. It finally returns to the Strategy and Outcome Space Exploration level the measure of competence progress for reaching $\tau_g$ or $\tau_d$.
\end{itemize}

The exploration in the three levels is the key to the robustness of SGIM-ACTS  in high dimensional policy spaces.

 \subsection{Policy Space Exploration}
\label {Pol}

 \subsubsection{Mimicry}
 \label{SectionMimic}
This function tries to mimic a demonstration $(\zeta_d, \tau_d)$  with policy parameters  $\theta_{im} = \theta_{d} + \theta_{rand} $ with a random movement parameter variation $|\theta_{rand}|< \epsilon$ and $\pi_{\theta_d}$ is the closest policy to reproduce $\zeta_d$. ${\theta_d}$ is computed by minimising over $\theta$ the distance between $\zeta_d$ and the motor primitives $\pi_{\theta}$. This function thus makes an estimate of $J(\tau_d,M(\theta))$ in the locality of $\theta_d$.  After a short fixed number of times,  SGIM-ACTS computes its competence at reaching the goal $\tau_{d}$. %(cf. eq. \ref{competence}).

\subsubsection{Goal-Directed Policy Optimisation}
\label{GoalDirectedPolicyOptimisation}
This  function  searches for policies $\pi_\theta$ that guide the system toward the goal $\tau_g$  by 1) building local models of $J$ during exploration that can be re-used for later goals and 2) updating its estimated inverse model $L^{-1}$. 
In the experiments below, exploration mixes local optimisation with the Nelder-Mead simplex algorithm \cite{Lagarias1998SJO} and global random exploration to avoid local minima. The measures are used to build memory-based local direct and inverse models,  using interpolation and more specifically locally weighted learning with a gaussian kernel such as presented in \cite{Atkeson1997R}.

\subsection{Strategy and Outcome Space Exploration}
\label{StratTask}

\subsubsection{Emulation}
In the emulation strategy, the learner explores outcomes $\tau_d$ that he observed from the demonstrations: $\tau_g \gets \tau_d$. The learner tries to achieve $\tau_d$ by goal-oriented policy optimisation, which allows data collection and updating of $L^{-1}$.

\subsubsection{Outcome and Strategy Interest Mapping}
$T$ is partitioned according to interest levels. We note  $\mathcal{R } =\{ R_i,  T = \cup_i R_i\}$ a partition of $T$. For each outcome $\tau$ explored with strategy $\sigma$, the learner  evaluates its competence progress, where competence measure assesses how close it can reach $\tau$:
$\gamma =  J(\tau,M(L^{-1}(\tau)))$.
A high value of $\gamma$ means a good competence at reaching the goal $y_g$  by strategy $\sigma$. 

For each episode, it  can compute its competence for the goal outcome at the beginning of the episode $\gamma_1$ and the end of the episode $\gamma_2$ after trying $nbA$ movements and measure its competence progress:

\begin{center}
\vspace{-0.8cm}
%\scriptsize
\begin{eqnarray}
prog =  2( sig(\alpha_p* \frac{\gamma_1 -\gamma_2}{|T_i| \cdot nbA} )-1 )   \text{ with } sig(x) = \frac{e^x + e^{-x}}{2}
\end{eqnarray}
\vspace{-0.6cm}
\end{center}

where $\alpha_p$ is a constant and $|T_i|$ is the size of the subspace $T_i$.

$T$ is partitioned so as to maximally discriminate areas according to their  competence progress, as described in Algorithm \ref{alg:UpdateRegions} and \cite{Baranes2013RAS}. For each strategy $\sigma$, we define a cost $\kappa (\sigma)$,  which are weights for the computation of the interest of each region of the outcome space. $\kappa (\sigma)$ represents the preference of the teachers to help the robot or not, or the cost in time and energy ... of each strategy, and in this study $\kappa(\sigma)$ are set to arbitrary constant values.   

We compute the  interest as  \textit{ the local competence progress, over a sliding time window of the $\mathbf{\delta}$ most recent goals attempted inside ${R}_i$ with strategy $\sigma$} which builds the list of competence progress measures $R_i(\sigma) = \{ progress_1,...progress_{|R_i(\sigma)| }\}$:

\begin{center}
\vspace{-0.8cm}
%\scriptsize
\begin{eqnarray}
interest_{R_i}(\sigma) =   \frac{\displaystyle mean_{j=|R_i(\sigma)|-\delta}^{|R_i(\sigma)|} progress_{j}}{ \kappa (\sigma) }
\label{interest}
\end{eqnarray}
\vspace{-0.6cm}
\end{center}

The partition of $  T$ is done recursively and so as to maximally discriminate areas according to their levels of interest. A split is triggered once a number of outcomes $g_{max}$ has been attempted inside $R_n$ with the same strategy $\sigma$. The split separates areas of different interest levels and different reaching difficulties. The split of a region $R_n$ into $R_{n+1}$ and $R_{n+2}$ is done by selecting among m randomly generated splits, a split dimension $j \in |T|$ and then a position $v_j$ (we suppose that $R_{n}\subset T_i \subset T $ with $T_i$ a n-dimensional space) such that:
\begin{itemize}
\item All the $\tau \in R_{n+1}$ have a jth component smaller than vj;
\item All the $\tau \in R_{n+2}$ have a jth component higher than vj;
\item It maximises the quantity	$Qual(j,vj)	= |R_{n+1}|. |R_{n+2}|$ $|interest_{R_{n+1}((\sigma))}	-	interest_{R_{n+2}}(\sigma) |$, where $|R_i|$ is  the size of the region $R_i$;
\end{itemize}

\begin{figure}
\input{PseudoCodeUpdateRegions.tex}
\vspace{-1cm}
\end{figure}

\subsubsection{Select Goal Outcome and Strategy}
\label{SectionSelectTaskStategy}

\begin{figure*}
\centering
\includegraphics[width=0.7\textwidth]{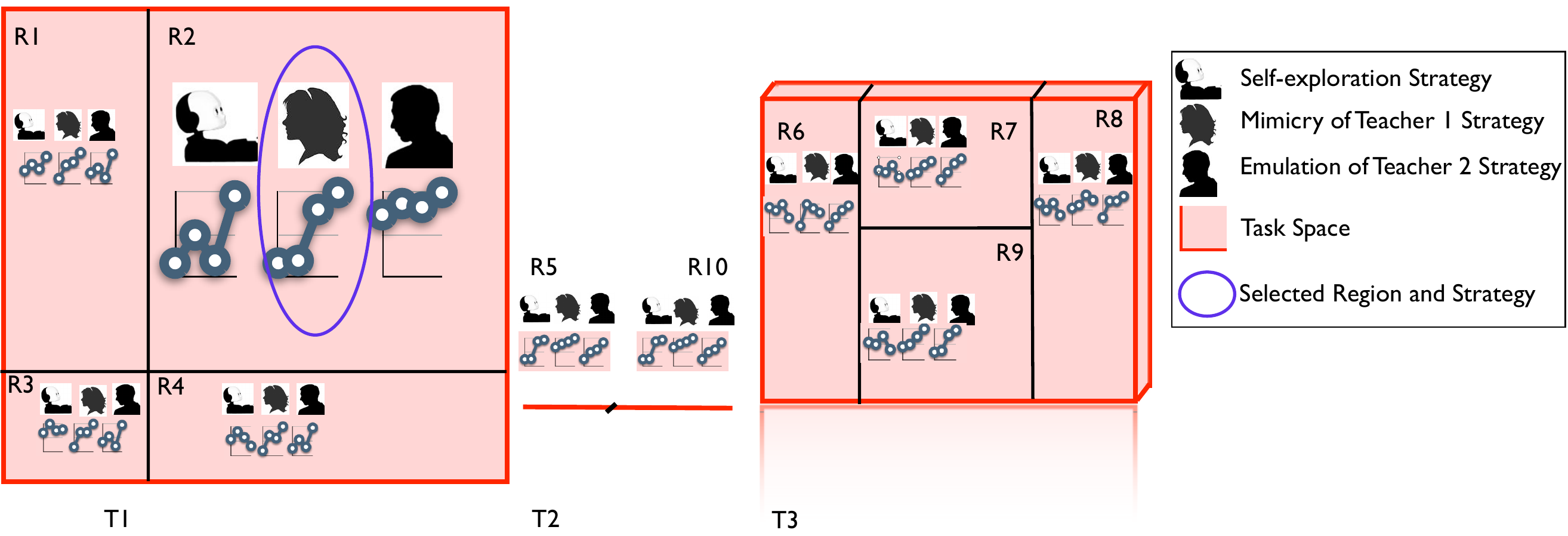}
\vspace{-0.5cm}
\caption{The selection of outcome and strategy is based on a partition of the outcome space with respect to different competence progress levels. We illustrate with the case of an outcome space of 3 different types of outcomes. $T = T1 \cup T2 \cup T3$ where $T1 \subset \mathbb{R}^2$, $T2 \subset  \mathbb{R}$ and $T3 \subset  \mathbb{R}^3$. T is partitioned in regions $R_i$ to which are associated measures of competences  $\gamma$ for each strategy. The "Select  Goal Outcome and Strategy" function chooses the (region, strategy) pair that makes the most competence progress.}
\vspace{-0.5cm}
\label{SelectTaskStategy}
\end{figure*}

In order to balance exploitation and exploration, the next goal outcome and strategy are selected according to one of the 3 modes, chosen stochastically with respectively probabilities p1, p2 and p3:

\begin{itemize}

\item mode 1: choose $\sigma$ and $\tau \in T$ randomly. It ensures a minimum of exploration of the full strategy and outcome spaces.

\item mode 2: choose the region $R_n(\sigma$) and thus  the strategy $\sigma$  with a probability proportional to its interest value $interest_{R_n}(\sigma$):
\vspace{-0.4cm}
\begin{eqnarray}
P_n(\sigma) = \frac{ interest_{R_n}(\sigma) - \textbf{min}(interest_{R_i})}{\sum_{i=1}^{|R_n|}interest_{R_i}(\sigma) - \textbf{min}(interest_{R_i})}
 \label{goalSelection}
\end{eqnarray}
\vspace{-0.4cm}

 A outcome $\tau$ is then generated randomly inside $R_n$. This mode uses exploitation to choose the region with highest interest measure.

\item mode 3: the strategy and regions are selected like in mode 2, but the outcome $\tau \in R_n$ is generated close to the already experimented one which received the lowest competence estimation.  This mode also uses exploitation to choose the best outcome and strategy with respect to interest measures.
\end{itemize}

 We illustrate in the following section this hierarchical algorithm through an illustration example where a robot learns to throw a ball or to place it at different angles with 7 strategies: intrinsically motivated exploration, mimicry from 3 teachers and emulation from 3 teachers.

\section{Throwing and Placing a Ball}
\label{Exp}

\subsection{Experimental Setup}

In our  simulated experimental setup, we have a 1 degree-of-freedom arm  place  a ball at different angles or throw the ball by controlling its angular acceleration $\ddot{\phi}$ (fig. \ref{ExperimentalSetup}).  The time evolution of its angular acceleration is described with motor primitives determined by 14 parameters. $\Pi \subset \mathbb{R}^{14}$ as described in  \ref{PolicyParameterSpace}. The outcome space is composed of 2 types of outcomes $T = T1 \cup T2$, that we detail in \ref{ThrowingOutcomes} and \ref{PlacingOutcomes}. 

\subsubsection{Policy Parameter Space}
\label{PolicyParameterSpace}
Starting from angle $\phi = 0$, the robot can control its angular acceleration $\ddot{\phi}$. Its movement is parameterised by ($ \ddot{\phi}_1, t_1, ...  \ddot{\phi}_7, t_7$) which defines the acceleration of the arm for the 7 durations $t_i$. It thus defines $\ddot{\phi}(t)$ as a piecewise constant function.  The policy parameter space is arbitrarily set to a 14 dimensional space.

\subsubsection{Throwing Outcomes}
\label{ThrowingOutcomes}
The first type of outcomes is the different distance x and height h at which the ball B  can be thrown.  $T1 = \{(x,h)\}$ is a continuous space of dimension 2.  The ball, initially in the robot's hand is first accelerated by the robot arm, and then automatically released: 
\begin{itemize}
\item at position $ \vec{OB}_{t=0}$ which is the position of the tip of the arm,  
\item with velocity  $\frac{d \vec{OB}}{dt}_{t=0}$  which magnitude is the velocity of the arm, and which direction is the tangent of the arm movement.
\end{itemize}
Then, the ball falls under gravity force, described by the equation:
\begin{equation}
\vec{OB}_t = \frac{\vec{g}}{2} \cdot t^2 + \frac{d \vec{OB}}{dt}_{t=0} \cdot t  +  \vec{OB}_{t=0},
\label{fall}
\end{equation}

where $\vec{g} $ is the gravity force.
x is therefore computed for $t_{impact}$, the time when the ball touches the ground,  or in other words the solution to the 2nd polynomial equation:
\begin{equation}
 \frac{-g}{2}\cdot t^2 + \frac{dz}{dt}_{t=0} \cdot t  +  z_{t=0} = 0
\end{equation}

The maximum height is also directly computed by equation:
\begin{equation}
h = z_{t=0} + \frac{(\frac{d {OB}}{dt}_{t=0})^2}{2g};
\end{equation}

To make the throwing less trivial, we also added a wall as an obstacle at x= 10. The ball can bounce on the wall using an immobile wall model and elastic collision.

\subsubsection{Placing Outcomes}
\label{PlacingOutcomes}
The second type of outcomes is placing a ball at different angles $\phi$. Therefore $T2$ is of dimension 1. To achieve an outcome in $T2$, the robot has to stop its arm in a direction  $\phi$ before releasing the ball, i.e. it learns to reach $\phi$ at a small velocity $|v|< |v_{max}|$.  

 Any policy would move the arm to a final angle $\phi$, but to "place" the ball at an angle, it also needs to reach a  velocity smaller than  $|v_{max}|$. Therefore placing a ball is difficult.

The robot learns which arm movement it needs to perform to either place at a given angle $\phi$ or to throw a ball at a given height and distance. Mathematically speaking, it learns highly redundant mappings between a 14-dimensional policy space and  a union of a 1D and a 2D continuous outcome spaces.

In our experimental setup, the outcome space is thus the union of two continuous spaces of different dimensionalities, related to throwing and placing skills, which makes it complex because of the continuous and composite nature of the space.
The complexity of the placing of the ball depends on the physics of the body and on the structure of motor commands. We choose to control the robot by angular acceleration to emphasise the difference in the ease of control between the "throwing outcomes" which require rather a velocity control, and the "placing outcomes" which require rather a position control. Given the motor control by acceleration and the encoding of motor primitives,  the placing outcomes are thus more difficult to achieve than the throwing outcomes.

\subsection{Several Teachers and Strategies}
We create simulated teachers by building 3 demonstration sets from which to pick a random demonstration when asked by the learner : 
\begin{itemize}
\item teacher 1 has learned how to throw a ball with SAGG-RIAC. The teacher 1 has the same motor primitives encoding as the learner, and the robot observes from the demonstrated trajectories directly the demonstrated  ($ \ddot{\phi}_1, t_1, ...  \ddot{\phi}_7, t_7$).

\item teacher 2 is an expert in placing, programmed by an explicit equation to place at any angle with a null velocity. The teacher 2 too has the same motor primitives encoding as the learner, and the robot observes from the demonstrated trajectories directly the demonstrated  ($ \ddot{\phi}_1, t_1, ...  \ddot{\phi}_7, t_7$).

\item teacher 3 is an expert in placing, except that in this case the learner faces correspondence problems and misinterprets the two parameters $\ddot{\phi}_6$ and $\ddot{\phi}_7$ as the opposite values.  In this experiment, we do not attempt to solve this correspondence problem. We also note that while the learner has issues mimicking teacher 3,  he has no issues emulating teacher 3, as the outcome space parametrisation is the same.
\end{itemize}

Therefore in our experiment,  the interactive learner can choose between 7  strategies : SAGG-RIAC autonomous exploration,  emulation of each of the 3 teachers or mimicry of each of the 3 teachers.

\subsection{Comparison of Learning Algorithms}

\begin{figure}
\centering
\includegraphics[width= 0.5\textwidth]{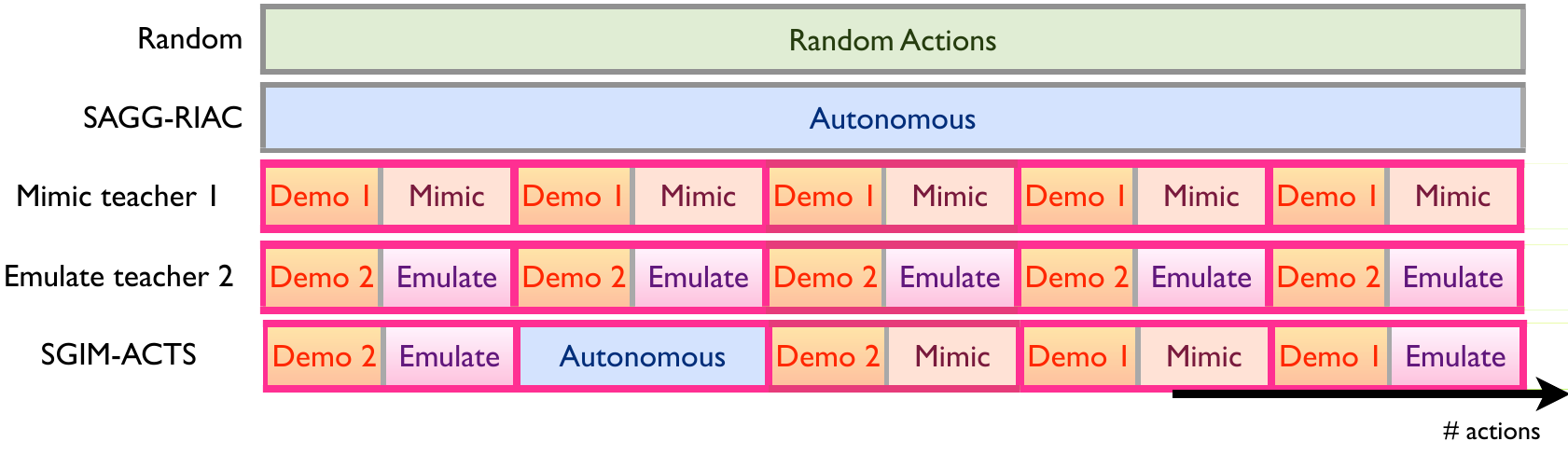}
\vspace{-1cm}
\caption{ Comparison of several learning algorithms %: Random exploration of the action space A, autonomous exploration SAGG-RIAC, Learning from Observation, Imitation learning and SGIM-D. The comparison is made through the same experimental duration (5000 actions performed by the robot), through the same teaching frequency (every 30 actions) and through regular evaluation (every 1000 actions). \\
}
\vspace{-0.6cm}
\label{ExperimentalProtocol}
%\vspace{-0.6cm}
\end{figure}

To assess the efficiency of SGIM-ACTS, we decide to compare the performance of several learning algorithms (fig. \ref{ExperimentalProtocol}):
\begin{itemize}
\item Random exploration : throughout the experiment, the robot learns by picking policy parameters randomly. It explores randomly the policy parameter space $\Pi$.
\item SAGG-RIAC : throughout the experiment, the robot uses active goal-babbling to explore autonomously, without taking into account any demonstration by the teacher, and is  driven by intrinsic motivation.
\item mimicry :  at a regular frequency, the learner  determines a goal $\tau_g$ where learning progress is maximal, and requests to the chosen teacher a demonstration. The teacher selects among his data set a demonstration $[\zeta_d, \tau_d]$ so that $\tau_d = argmin_{\tau \in \{Demo Set\}} ||\tau_g - \tau|| $. The learner mimics the demonstrated policy $\zeta_d$ by repeating the movement with small variations.
\item emulation : at a regular frequency, the learner  determines a goal $\tau_g$ where learning progress is maximal, and requests to the chosen teacher a demonstration. The teacher selects among his data set a demonstration $[\zeta_d, \tau_d]$ so that $\tau_d = argmin_{\tau \in \{Demo Set\}} ||\tau_g - \tau|| $. The learner tries to reproduce the outcome $\tau_d$.
\item SGIM-ACTS : interactive learning where the robot learns by actively choosing between intrinsic motivation strategy or one of the social learning strategies with the chosen teacher: mimicking or emulation.
\end{itemize}

We run simulations with the following parameters. The costs of all socially guided strategies $\kappa (\sigma)$ are set to $2$, and the cost of intrinsic motivation is set to 1. The probabilities for the different modes of selecting a region of the outcome space and a strategy are: p1 = 0.05, p2 = 0.7 and p3 = 0.25. Other parameters are  $\epsilon = 0.05, g_{max} = 10$, $\alpha_p = 1000$ and $v_{max} = 0.01$.

 For each experiment, we let the robot perform 8000 actions in total, and evaluate its performance every 1000 actions, by requiring the system to produce outcomes from a benchmark set that is evenly distributed in the outcome space and independent from the learning data.

\subsection{Results}

\begin{figure}
\centering
\includegraphics[width=0.5 \textwidth]{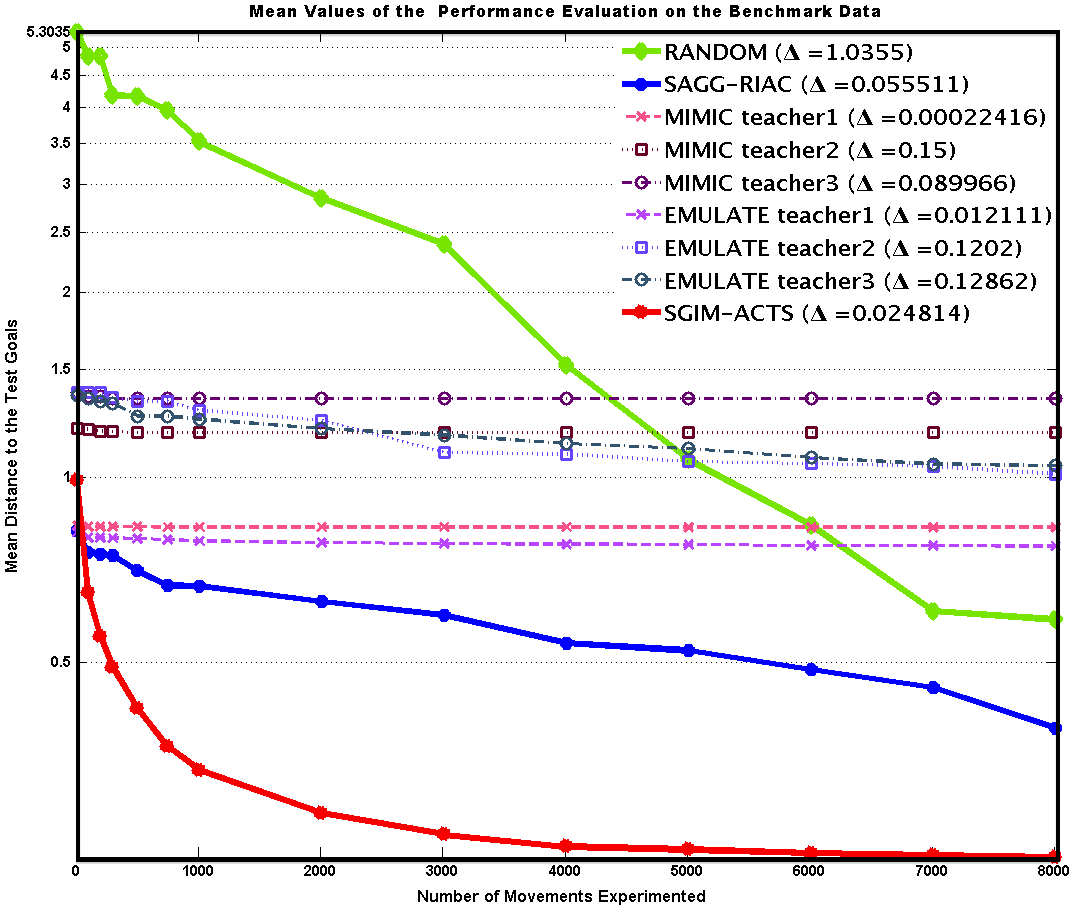}
\vspace{-0.8cm}
\caption{ Mean error for the different learning algorithms  averaged over the two sub outcome spaces (final variance value $\Delta$ is indicated in the legend) . 
}
\vspace{-0.6cm}
\label{error}
%\vspace{-0.6cm}
\end{figure}

\begin{figure}
\centering
\includegraphics[width= 0.5\textwidth]{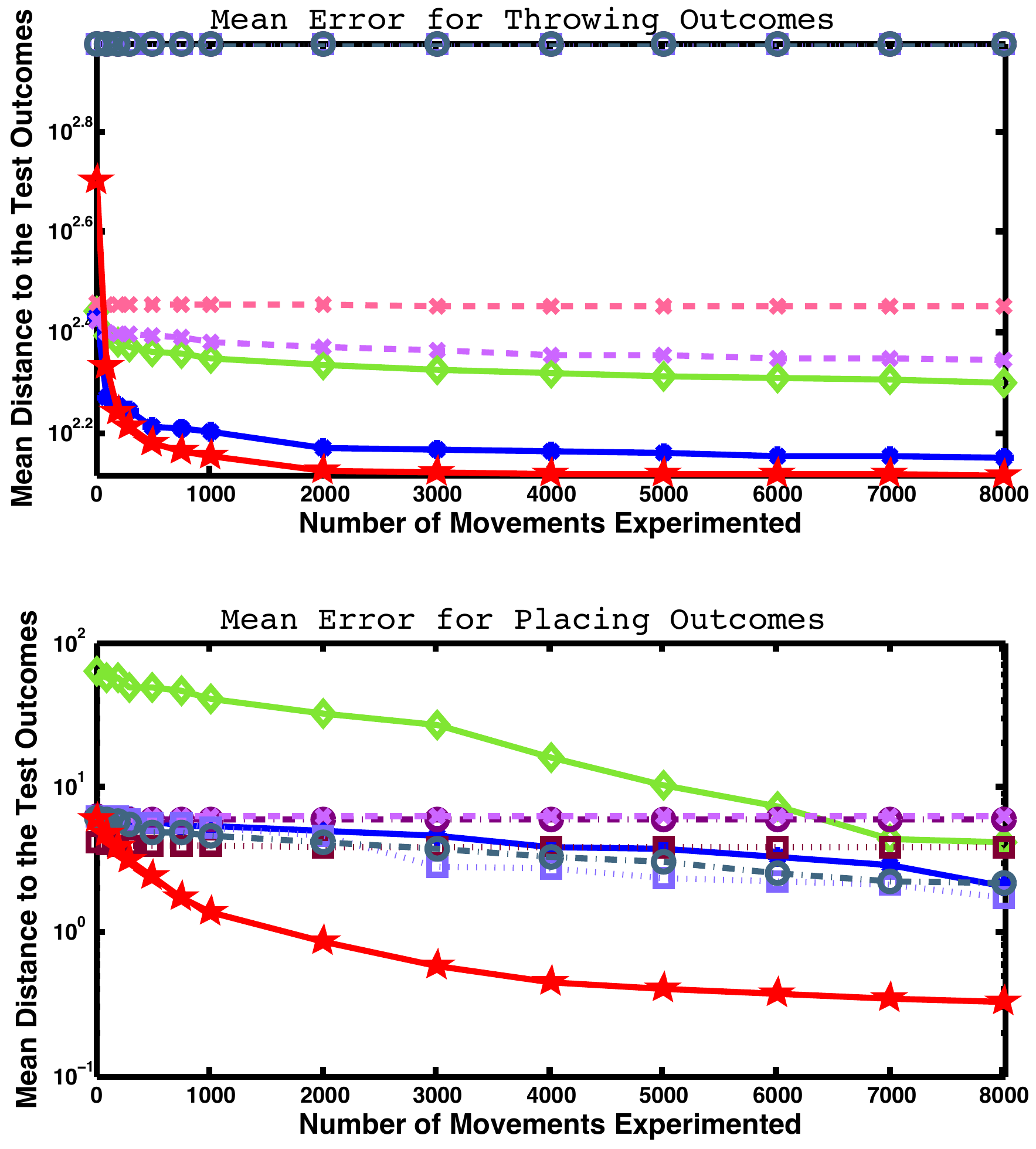}
\vspace{-0.8cm}
\caption{ Mean error for the different learning algorithms for each of the throwing outcomes and placing outcomes separately. The legend is the same as in fig. \ref{error}. }
\vspace{-0.6cm}
\label{errorPerTask}
%\vspace{-0.6cm}
\end{figure}

\begin{figure}
\centering
\includegraphics[width= 0.5\textwidth]{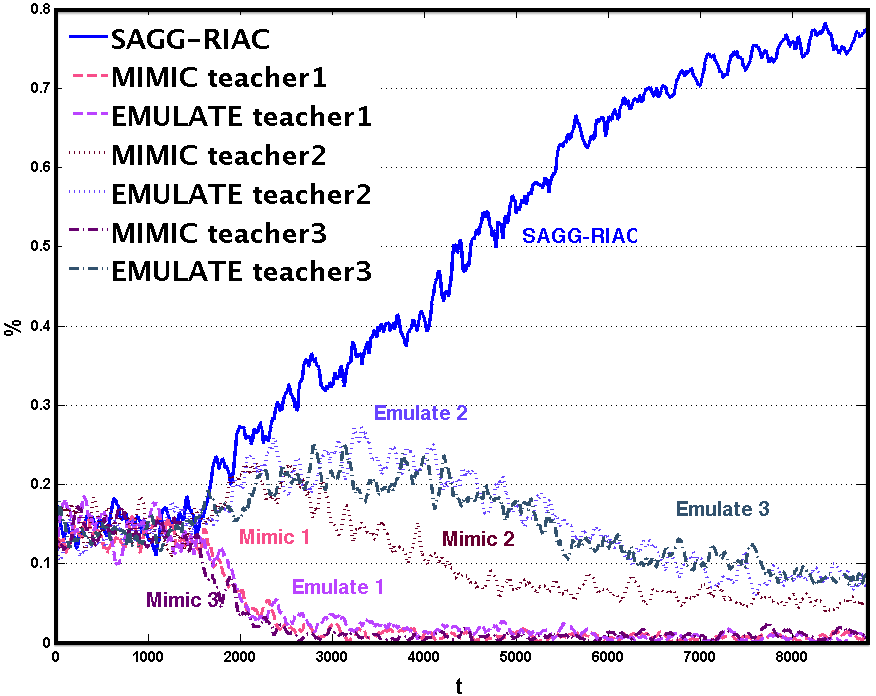}
\vspace{-0.8cm}
\caption{ Strategy chosen by SGIM-ACTS through time: percentage of times each strategy is chosen for several runs of the experiment.}
\vspace{-0.6cm}
\label{strat}
%\vspace{-0.6cm}
\end{figure}

\begin{figure}
\centering
\includegraphics[width= 0.4\textwidth]{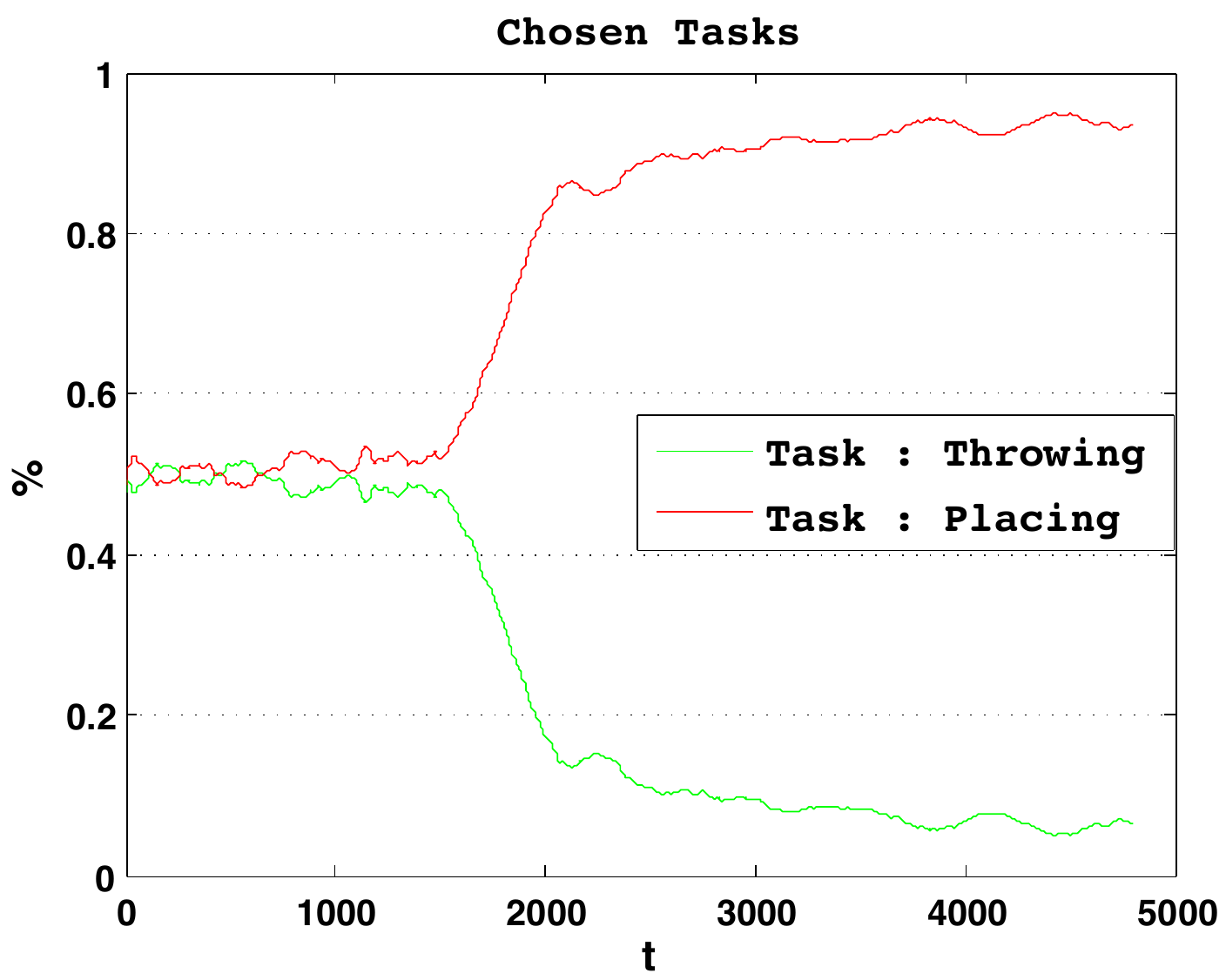}
\vspace{-0.6cm}
\caption{ Outcome chosen by SGIM-ACTS through time: percentage of times each kind of outcome is chosen for several runs of the experiment.}
\vspace{-0.6cm}
\label{outcome}
%\vspace{-0.6cm}
\end{figure}

\begin{figure}
\centering
\includegraphics[width=0.45 \textwidth]{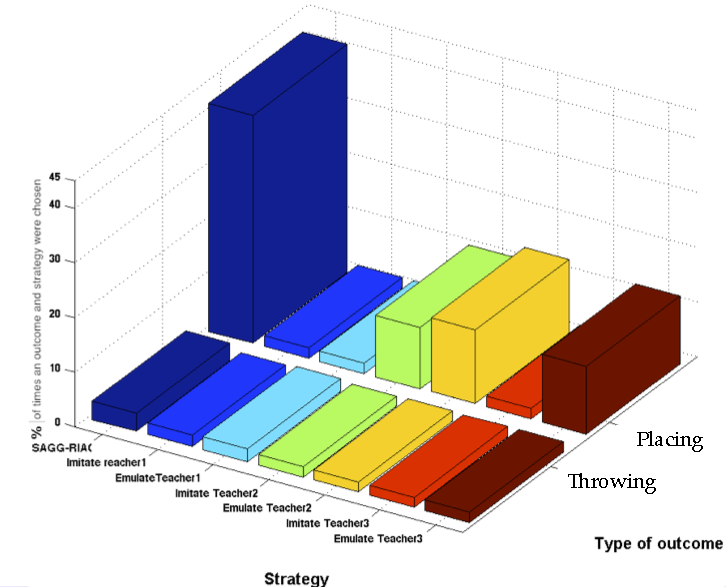}
\vspace{-0.6cm}
\caption{Consistency in the choice of outcome, teacher and strategy: percentage of times each strategy, teacher and outcome are chosen over all the history of the robot. }
\vspace{-0.6cm}
\label{concord}
%\vspace{-0.6cm}
\end{figure}

The comparison of these four learning algorithms in fig. \ref{error} shows that SGIM-ACTS decreases its cumulative error for both placing and throwing. It performs better than autonomous exploration by random search or intrinsic motivation, and better than any socially guided exploration with any teacher.  Fig. \ref{errorPerTask} details that SGIM-ACTS error rate for both placing  and throwing is low. For throwing, SGIM-ACTS performs the best in terms of error rate and speed because it could find the right strategy.  We also note that random exploration and SAGG-RIAC also perform well for solving the 2nd degree polynomial equation (\ref{fall}) to achieve throwing outcomes. While mimicking and emulating teacher 1 decreases the error as expected, mimicking and emulating a teacher who is expert in another kind of outcomes and is bad in that outcome  leaves a high error rate. For placing, SGIM-ACTS makes less error than all other algorithms. Indeed, as we expected, mimicking the teacher 2, and emulating teachers 2 and 3 enhances low error rates, while mimicking a teacher with correspondence problem (teacher 3)  or an expert on another outcome (teacher 1) gives poor result. We also note that for both outcomes, mimicry does not lead to  important learning progress, and the error curve is almost flat. This is due to the lack of exploration which leads the learner to ask demonstrations for outcomes only in a small subspace.

Indeed,  we see in fig. \ref{strat} which illustrates the percentage times each strategy is chosen by SGIM-ACTS with respect to time, that mimicry of teacher 3, which lacks efficiency because of the correspondence problem, is seldom chosen by SGIM-ACTS. Mimicry and emulation of teacher 1 is also little used because autonomous learning learns quickly throwing outcomes. Teachers 2 and 3 are exactly the same with respect to the outcomes they demonstrate, and are emulated in the same proportion. This figure also shows that the more the learner cumulates knowledge, the more autonomous he grows : his percentage of autonomous learning increases steadily.

Not only does he choose the right strategies, but also the right outcome to concentrate on. Fig. \ref{outcome} shows that he concentrates in the end more on placing, which are more difficult.

Finally, fig. \ref{concord} shows the percentage of times over all the experiments where he chooses at the same time each outcome type, a strategy and a teacher. We can see that for the placing outcomes, he seldom requests help from the teacher 1, as he learns that teacher 1 does not know how to place the ball. Likewise, because of the correspondence problems, he does not mimic teacher 3. But he learns that mimicking teacher 2 and emulating teachers 2 and 3 are useful for placing outcomes. For the throwing outcomes, he uses slightly more the autonomous  exploration strategy, as he can learn efficiently by himself. The high percentage for the other strategies is due to the fact that the throwing outcomes are easy to learn, therefore are learned in the beginning when a lot of sampling of all possible strategies is carried out. SGIM-ACTS is therefore consistent in its choice of outcomes , data collection strategies and teachers.

\section{Conclusion and Discussion}
  
  We presented the \textbf{SGIM-ACTS} (Socially Guided Intrinsic Motivation with Active Choice of Teacher and Strategy) algorithm that efficiently and actively combines autonomous self-exploration and interactive learning, to address the learning of multiple outcomes, with outcomes of different types, and with different data collection strategies. In particular, it learns actively to decide on  the fundamental questions of programming by demonstration: \textit{what and how} to learn; but also \textit{what, how, when and who} to imitate.  This interactive learner decides efficiently and coherently whether to use social guidance. It learns when to ask for demonstration, what kind of demonstrations (action to mimic or outcome to emulate) and who to ask for demonstrations, among the available teachers. Its hierarchical architecture bears three levels. The lower level explores the policy parameters space to build skills for determined goal outcomes. The upper level explores the outcome space to evaluate for which outcomes he makes the best progress. A meta-level actively chooses the outcome and data collection strategy that leads to the best competence progress. We showed through our illustration example that SGIM-ACTS can focus on the outcome where it learns the most, while choosing the most appropriate associated data collection strategy. The active learner can explore efficiently a composite and continuous outcome space to be able to generalise for new outcomes of the outcome spaces. 

SGIM-ACTS has been shown an efficient method for learning with multiple teachers and multiple outcome types. The number of outcomes used in the experiment is infinite, with a continuous outcome space that is made of 2 types of outcomes, but all the formalism and framework is in principle scalable to a higher number of types of outcomes. Likewise, the method should apply to domestic or industrial robots who usually interact with a finite number of teachers. Even in the case of correspondence problems, the system still takes advantage of the demonstrations to bias its exploration of the outcome space. When the discrepancies between the teacher and the learner are small, demonstrations advantageously bias the exploration of the outcome space, as argued in \cite{Nguyen20122IISRHIC}. Future work should test SGIM-ACTS on more complex environments, and with real physical robots and everyday human users.
It would also be interesting to compare the outcomes selected by our system to developmental behavioural studies, and highlight developmental trajectories.

\paragraph{Acknowledgement}
This work was supported by the French ANR program (ANR 2010 BLAN 0216 01) through Project MACSi, as well by ERC Starting Grant EXPLORERS 240007.

%\begin{thebibliography}{99}
\bibliography{flowers}
\bibliographystyle{plain}
%\end{thebibliography}

\end{document}